\documentclass[conference]{IEEEtran}
\IEEEoverridecommandlockouts
\usepackage{cite}
\usepackage{amsmath,amssymb,amsfonts}
\usepackage{algorithmic}
\usepackage{graphicx}
\usepackage{textcomp}
\usepackage{xcolor}
\usepackage{hyperref}
\usepackage{caption}
\def\BibTeX{{\rm B\kern-.05em{\sc i\kern-.025em b}\kern-.08em
    T\kern-.1667em\lower.7ex\hbox{E}\kern-.125emX}}
\begin{document}

\title{A Dynamic Weighting Strategy to Mitigate Worker Node Failure in Distributed Deep Learning}

\author{\IEEEauthorblockN{Yuesheng Xu}
\IEEEauthorblockA{\textit{Computer Science \& Engineering} \\
\textit{Lehigh University}\\
Bethlehem, USA \\
yux220@lehigh.edu}
\and
\IEEEauthorblockN{Arielle Carr}
\IEEEauthorblockA{\textit{Computer Science \& Engineering} \\
\textit{Lehigh University}\\
Bethlehem, USA \\
arg318@lehigh.edu}

}

\maketitle

\begin{abstract}
The increasing complexity of deep learning models and the demand for processing vast amounts of data make the utilization of large-scale distributed systems for efficient training essential. These systems, however, face significant challenges such as communication overhead, hardware limitations, and node failure. This paper investigates various optimization techniques in distributed deep learning, including Elastic Averaging SGD (EASGD) and the second-order method AdaHessian. We propose a dynamic weighting strategy to mitigate the problem of straggler nodes due to failure, enhancing the performance and efficiency of the overall training process. We conduct experiments with different numbers of workers and communication periods to demonstrate improved convergence rates and test performance using our strategy.
\end{abstract}


\section{Introduction}
In recent years, deep learning has proven to be the most successful tool in many critical fields from computer vision to natural language processing\cite{chai2021deep}. However, the increasing complexity of deep learning models and the exponential growth in data render large-scale distributed settings essential in order to effectively manage computational loads and speed up the training process. At the same time, large-scale distributed deep learning also introduces a new set of challenges. These include hardware limitations, computational complexity, communication costs, data management, and the need to improve convergence speed, all of which are critical to optimize the efficient deployment of advanced deep learning models.

In this paper, we conduct an in-depth study of two optimization techniques, Elastic Averaging Stochastic Gradient Descent (EASGD)\cite{zhang2015deep} and the second-order method AdaHessian\cite{yao2021adahessian}. By combining and modifying these methods, we develop a dynamic weighting strategy aimed to mitigate the impact of straggler nodes due to failures and enhance both the performance and efficiency of the distributed training process.

The remainder of this paper is organized as follows: Section II describes the problem setting and the assumptions made in our approach. Section III discusses types of parallelism in distributed deep learning. Section IV reviews related work. Section V presents our proposed method, including the data overlap and dynamic weighting strategy. Sections VI and VII describe the experimental settings and present the results, respectively. Finally, in Section VIII we provide conclusions and suggestions for future work.

\section{Problem Setting}
We consider a deep neural network \( f_{\theta} \) parameterized by weight vector \( \theta \), where the network is defined as a function \( f_{\theta}: \mathbb{R}^{d} \to \mathbb{R} \) mapping an input space \( \mathbb{R}^{d} \) to an output space \( \mathbb{R} \) for a general classification task. The goal is to find the optimal parameters \( \theta^* \) that minimize the loss function \( \mathcal{L}(x_i, y_i;\theta) \) given the dataset \( \mathcal{D} = \{(x_i, y_i)\}_{i=1}^{n} \), where $(x_i, y_i) $ are the inputs and corresponding labels. The optimization problem can be formally stated as:

\begin{equation}
\theta^* = \arg \min_{\theta} \mathcal{L}(x_i,y_i;\theta).
\end{equation}

In practice, we have large datasets and complex models with millions -- and potentially more -- of parameters. Instead of using a full batch optimization method\cite{bottou2010large,ruder2016overview}, we instead perform updates on the model parameters using mini-batches\cite{bottou2010large,ruder2016overview}. This is formalized as follows:

\begin{equation}
g_t \leftarrow \frac{1}{m} \sum_{i}  \nabla\mathcal{L}(x_i, y_i;\theta),
\end{equation}

\begin{equation}
\theta_{t+1} \leftarrow \theta_{t} - \eta_t g_t,
\end{equation}
where \( g_t \) is the estimated gradient computed using the mini-batch of \( m \) samples and \( \eta_t\) is the learning rate at the \( t \)-th iteration.

The most commonly used optimization algorithms are first-order methods such as stochastic gradient descent (SGD)\cite{bottou1998online}, Momentum\cite{nesterov2005smooth,sutskever2013importance}, and ADAM\cite{kingma2014adam}. There are also second-order methods that impose the update rule:
\begin{equation}
\theta_{t+1} \leftarrow \theta_t - \eta_t D_t g_t,
\end{equation}
where $D_t$ is an approximation of the inverse of the Hessian matrix or any diagonal matrix that captures curvature information\cite{broyden1970class,nocedal2006numerical,dennis1996numerical}.

Our approach is based on the following assumptions:

\begin{itemize}
    \item We consider only asynchronous distributed training procedures.
    \item There is a single master node and multiple worker nodes, each initially with a copy of the master's model.
    \item We assume a similar data distribution at each worker node.
    \item The communication cost between the master and worker nodes dominates the computation cost of a single worker node.
\end{itemize}
Of course, worker nodes can fail. In this paper, and operating under the assumptions listed above, we aim to handle worker node failure at the algorithmic level. Implementation of hardware level or network level mechanisms to detect the absence of a node is future work.
\section{Parallelism in Distributed Deep Learning}
\subsection{Model Parallelism}
\begin{figure}[t]
    \centering
    \includegraphics[width=0.5\textwidth]{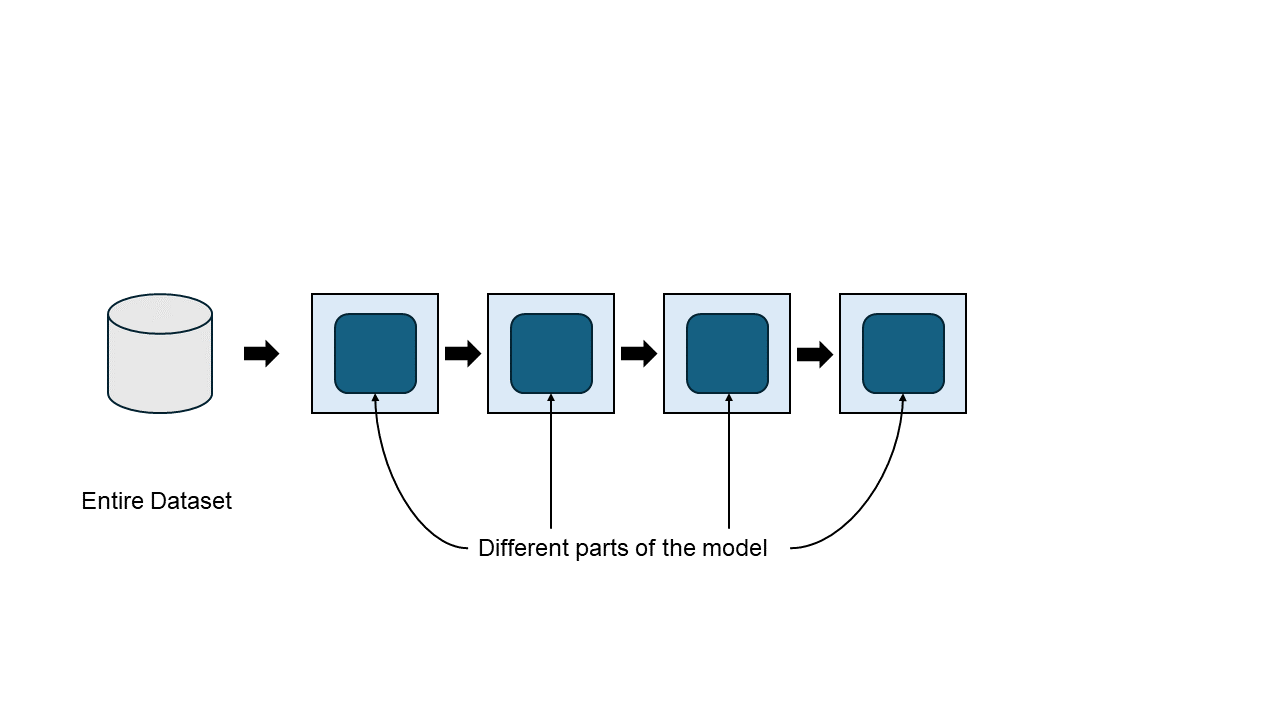}
    \caption{This diagram illustrates the concept of \textbf{model parallelism}. Each worker holds a different segment of the model, allowing for parallel computation and handling of very large models that cannot fit into a single device.}
    \label{fig:model}
\end{figure}

Model parallelism refers to the distribution of a neural network's architecture across different computational units\cite{dean2012large}. In this approach, different parts of the model are located on different devices, allowing for the simultaneous computation of different layers or segments of a network. See figure \ref{fig:model} for a visual representation.

For instance, let \( f_{\theta_1} \) and \( f_{\theta_2} \) be two distinct parts of a neural network model that can be trained in parallel on separate devices, where the output of $f_{\theta_1}$ serves as the input of $f_{\theta_2}$:

\begin{equation}
f_{\theta}(x) = f_{\theta_2}(f_{\theta_1}(x)).
\end{equation}
This can be a solution for training extremely large models that do not fit into the memory of a single device.
However, this approach creates a dependency chain as subsequent computations must wait for the preceding ones to complete before proceeding. 
\begin{figure}[t]
    \centering
    \includegraphics[width=0.5\textwidth]{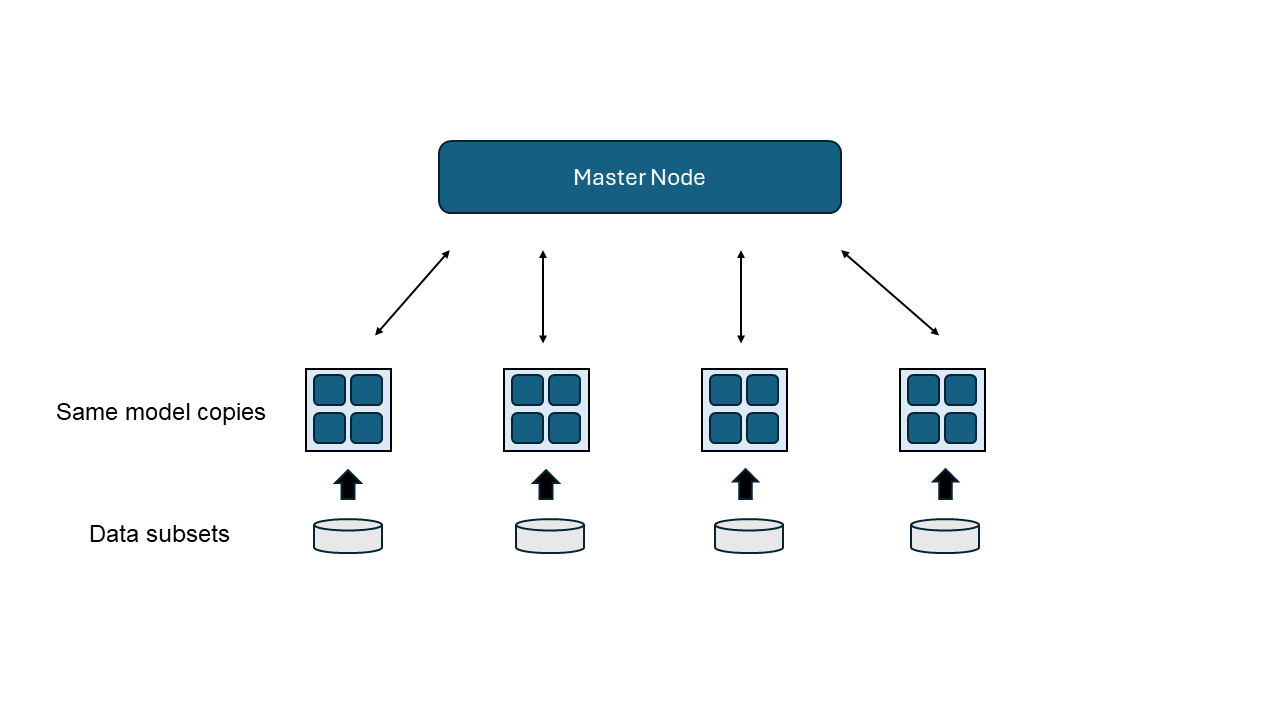}
    \caption{This diagram illustrates the concept of \textbf{data parallelism}. Each worker node holds a subset of the dataset and communicates with the master node to update both its own model as well as the
aggregated model. }
    \label{fig:data}
\end{figure}

\subsection{Data Parallelism}
In contrast to model parallelism, data parallelism splits the dataset into smaller subsets and distributes the subsets across multiple worker nodes; see figure \ref{fig:data}. Each worker node processes its own branch of data and updates its local model parameters independently. At each iteration \(t\), given a local subset of data \(\mathcal{D}_t^{i}\), the local loss for worker \(i\) is computed as follows:

\begin{equation}
\mathcal{L}^{i}(\theta^{i}_t) = \frac{1}{|\mathcal{D}_t^{i}|} \sum_{(x_j, y_j) \in \mathcal{D}_t^{i}} \mathcal{L}(x_j, y_j; \theta^{i}_t),
\end{equation}
where \(\theta^i_t\) represents the model parameters at iteration \(t\). Each worker node then communicates its local model parameters to the master node. The global parameter update at iteration \(t\) can be expressed in a more general form as:

\begin{equation}
\theta_{t+1} = \textbf{Agg}(\{\theta_t^{i}\}_{i=1}^{N}),
\end{equation}
where \(N\) is the total number of workers and \(\textbf{Agg}(\cdot)\) is a function that combines the local parameters from all workers. This step ensures that all workers contribute to the learning process and the aggregated model converges to a solution that is informed by the entire dataset.

\subsection{Synchronous and Asynchronous Methods}
In general, data-parallel distributed optimization methods are categorized into two distinct types: synchronous and asynchronous. In the synchronous approach, all worker nodes must synchronize their updates for each communication period, which can lead to bottlenecks if some nodes are slower than the others. The necessity for the master node to wait for all worker nodes makes the synchronous approach less scalable in a large-scale system\cite{chen2016revisiting}.

On the other hand, asynchronous methods do not require all nodes to synchronize their updates with the master node. A worker node can proceed to the next task after its communication with the master node without waiting for updates from the other worker nodes. The asynchronous approach leads to a faster overall computation time, especially with some advanced scheduling to avoid communication collision among worker nodes. However, there are some pitfalls for the asynchronous method: staleness of gradients and gradient noise\cite{chen2016revisiting}. The elastic averaging technique can mitigate the staleness by pulling worker and master together. Data overlap can reduce the noise both in gradients and Hessian diagonal approximation.

\section{Related Work}
\subsection{Elastic Averaging Stochastic Gradient Descent}
EASGD\cite{zhang2015deep} is an advanced optimization algorithm that improves the efficiency and robustness of distributed deep learning systems. At each iteration or communication period $t$, the update rules for one worker node $i$ and the master node $m$ are:
\begin{equation}
\theta_{t+1}^i = \theta_t^i  - \alpha (\theta_t^m - \theta_t^i),
\label{eq:worker_update}
\end{equation}
\begin{equation}
\theta_{t+1}^m = \theta_t^m + \alpha (\theta_t^m - \theta_t^i),
\label{eq:master_update}
\end{equation}
where $\alpha$ is a fixed ``moving rate" that controls the exploration and exploitation across worker nodes. It does so by regulating how much each worker node's parameters are influenced by the global parameters and vice versa. The parameters in the master node serve as an aggregated model. The hyper-parameter $\alpha$ can be seen as a force pulling worker node $i$'s parameters and master node's parameters towards each other. 

One problem, however, is if some worker node fails to synchronize with the master node. For the next possible communication which the failed worker node can successfully synchronize with the master node, the outdated model from the worker node is likely to cause adverse effects on the aggregated model. To mitigate the effect from the failed worker node, we can dynamically adjust the pulling force between worker and master nodes by employing a mechanism to identify these potential bad cases. If a worker node fails, we want the aggregated model to exert a larger pulling force on the ``bad" model while the ``bad" model exerts a smaller pulling force on the aggregated model. In doing so, the failed worker node has less of a (negative) impact on the overall aggregated model. We will discuss the details of handling the failed worker in the Section V.

\subsection{Second-Order Methods}
An optimizer plays a significant role in the model training. In general, there are two types of commonly used optimization methods: first-order and second-order. 

First-order methods primarily use the first derivatives of the loss function with respect to the model parameters. Stochastic Gradient Descent (SGD) is a well-known optimizer for its simplicity and effectiveness on a wide range of deep learning tasks. The primary advantage of first-order methods is their computational efficiency, which makes them particularly suitable for large-scale applications. They are less memory-intensive as they do not compute and store second-order information. However, their major disadvantage lies in their potentially slower convergence rates due to their reliance on local gradient information only, especially in the presence of ill-conditioned optimization landscapes or when navigating flat regions in the loss surface.

On the other hand, second-order methods use both first- and second-order information, where the second-order information is used to precondition the gradient. The main challenge associated with second-order methods stems from the computational complexity when computing the inverse of the Hessian matrix, which naively requires $O(n^3)$ operations for $n$ parameters. In practice, however, these methods always compute some approximation of the inverse of the Hessian, reducing complexity to $O(n)$. Moreover, in distributed systems, the dominant factor often shifts from computational cost to communication overhead. The frequent exchange of parameters between worker nodes and the master node can substantially outweigh the computational costs associated with updating each worker node's model. Therefore, it is beneficial for worker nodes to take slower yet accurate steps. If the increase in computational cost does not grow exponentially with the use of second-order methods, this strategy is highly likely to effectively reduce the overall wall-clock time by minimizing communication rounds while maintaining robust convergence.

\textbf{AdaHessian}\cite{yao2021adahessian} is a second-order optimizer that uses the approximated diagonal of the Hessian matrix to adaptively adjust the learning rate. We use this optimizer as the backbone of our distributed optimization system. There are three components that allow AdaHessian to effectively utilize second-order information. 

Firstly, it uses Hutchinson's method to approximate the Hessian diagonal\cite{bekas2007estimator}:

\[
\text{diag}(\mathbf{H}) \approx \frac{1}{n} \sum_{i=1}^{n} (\mathbf{z}_i \odot (\mathbf{H}\mathbf{z}_i)),
\]
where $\textbf{H}$ is the Hessian and $\mathbf{z}_i $ is a Rademacher vector. The Hessian vector product $\mathbf{H}\mathbf{z}_i$ takes the same amount of time as one back-propagation. Secondly, to reduce the Hessian variance, the spatial averaging Hessian diagonal for each parameter is computed by taking the average around its neighbours. Finally, \textbf{AdaHessian} adjusts the learning rate adaptively, similar to the ADAM optimizer\cite{kingma2014adam}. They differ in their calculation of the second moment: the
gradient is replaced by the spatial averaging Hessian diagonal.

\section{Method}
Our method builds on the concept of asynchronous averaging of the worker and master's parameters with the use of the second-order methods to address the challenge of \textbf{parameter desynchronization} caused by straggler nodes. Due to prior failures, such nodes update their parameters less frequently than other nodes. Previous work on using data encoding to mitigate the impact of straggler nodes by embedding redundancy directly in the data has been introduced in \cite{karakus2017straggler}.

\subsection{Data Overlap}
In distributed deep learning, particularly when using second-order optimization methods, the overlap of data across different worker nodes can be beneficial to both the performance as well as the stability of the training process. Consider the the set consisting of $n$ data points $\mathcal{D} = \{(x_i, y_i)\}_{i=1}^{n}$ and $k$ workers. We randomly sample a  of data points $\mathcal{O}$ of size $|\mathcal{O}|$. Ideally, $|\mathcal{O}|$ should be chosen carefully to balance between too much and too little overlap. For instance, if $|\mathcal{O}|$ is chosen to be too large, this diminishes the benefits of distributed system as it counteracts the purpose of speeding up the process by parallelizing the computation. Conversely, a small or no overlap can potentially lead to sub-optimal Hessian approximations. In more complex and realistic environments, the data distribution across different workers can vary significantly. Without a certain degree of data overlap, workers in general can encounter very different loss landscapes, thus generating Hessian approximations with high variance. Ongoing work focuses on addressing the performance of our method in real-world applications and scenarios.

All workers share a subset of \(|\mathcal{O}| = o\) data points. The remaining data points \(\mathcal{D} - \mathcal{O}\) are then randomly distributed among the \(k\) workers. Each worker \(w_j\) will receive the shared subset \(\mathcal{O}\) and a unique subset \(\mathcal{S}_j\) of size \(|\mathcal{S}_j| = \left\lfloor \frac{n - o}{k} \right\rfloor\). Therefore, the dataset assigned to worker \(w_j\) can be characterized as:
\[ \mathcal{D}_j = \mathcal{O} \cup \mathcal{S}_j, \]
where \(j \in \{1, 2, \ldots, k\}\), \(\bigcup_{j=1}^{k} \mathcal{S}_j = \mathcal{D} - \mathcal{O}\) and \(\mathcal{S}_i \cap \mathcal{S}_j = \emptyset\) for \(i \neq j\).

\begin{figure}[t]
    \centering
    \includegraphics[width=0.5\textwidth]{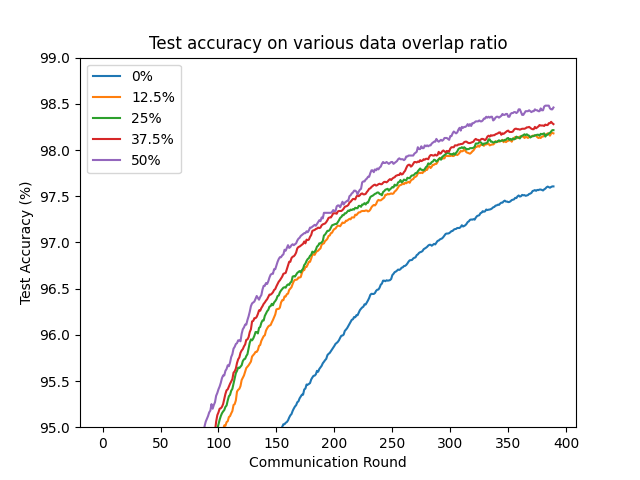}
    \captionsetup{font=scriptsize}
    \caption{Data overlap ratios: $\{0\%, 12.5\%, 25\%, 37.5\%, 50\%\}$}
    \label{fig:overlap}
\end{figure}

We provide empirical observations to justify the careful balancing of the overlap. Let \(r=\frac{o}{n}\) be the ratio of the overlapped data shown by all $k$ worker nodes. As shown in figure \ref{fig:overlap}, we compare different data overlap ratio on $\textbf{EAHES}$. As expected, we observe a positive relationship between the data overlap ratio and test accuracy. 
\begin{figure*}[h!]
    \centering
    \includegraphics[width=\textwidth]{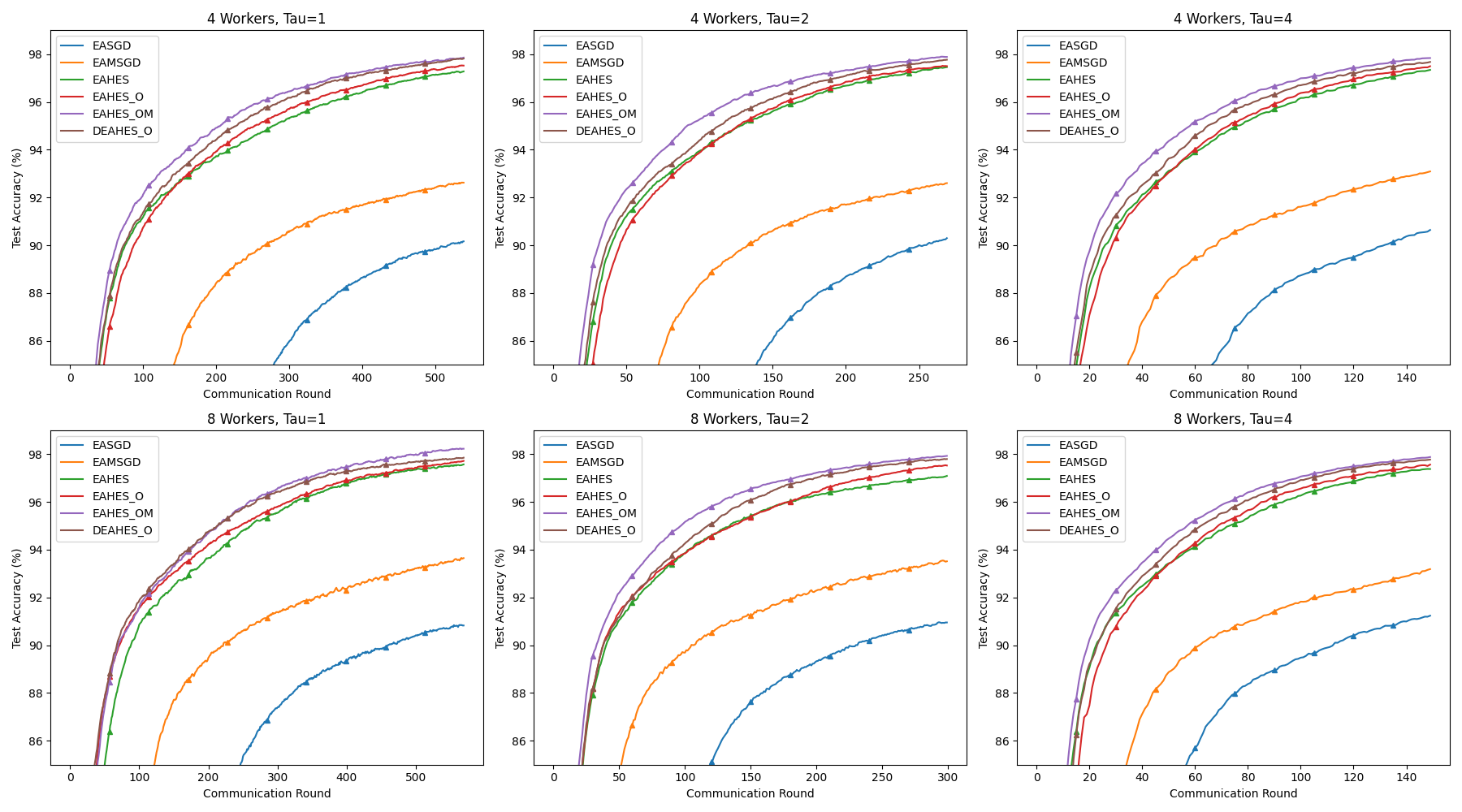}
    \caption{Test accuracy over communication rounds for workers $k \in \{4, 8\}$ and communication period $\tau \in \{1,2,4\}$. Each experiment is averaged over 3 runs.}
    \label{fig:accuracy}
\end{figure*}
\begin{figure*}[h!]
    \centering
    \includegraphics[width=\textwidth]{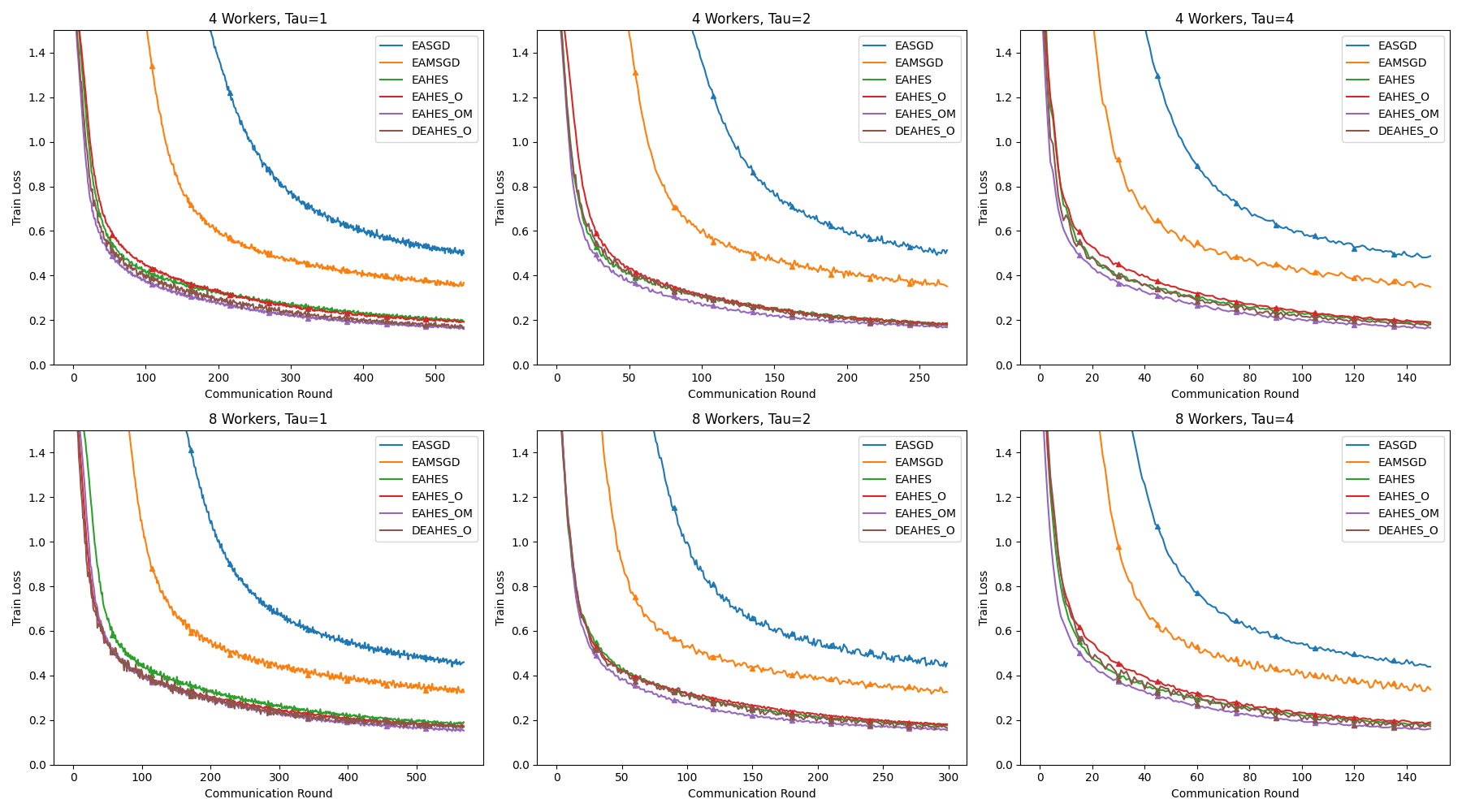}
    \caption{Training loss over communication rounds for workers $k \in \{4, 8\}$ and communication period $\tau \in \{1,2,4\}$. Each experiment is averaged over 3 runs.}
    \label{fig:loss}
\end{figure*}

\subsection{Dynamic Weight}
In order to mitigate the influence of a failed worker on the master node, we propose a dynamic weighting strategy. Intuitively, we want to (1) reduce the bad influence from the failed node by increasing $\alpha$ in equation (\ref{eq:worker_update}), and (2) correct the failed node using the aggregated model by decreasing $\alpha$ in equation (\ref{eq:master_update}). Instead of a fixed $\alpha$ value, we map a raw score that measures the change in distance between a worker and the master node to the dynamic weight. We choose piece-wise linear functions based on the observation that if a worker fails, its raw score becomes negative in the next few time steps. If a worker node works properly, its raw score usually is small and positive, and we approximate the behavior of EASGD in this case. For a worker node $w_i$, we compute the 2-norm of its model $\theta^i$ and the estimate of the master model $\tilde{\theta}^m$. In practice, we can acquire this estimation from other workers efficiently since communication among workers is much faster. We further define a term to estimate how far away worker model $w_i$ is from the aggregated model:
\begin{align*}
    u^i_t = \log (||\theta^i_t - \tilde{\theta}^m_t||).
\end{align*}
We store the $p$ most recent $u^i$ -- that is, $\{u^i_{t-p-1},...,u^i_{t-2},u^i_{t-1},u^i_{t}\}$ -- and compute a weighted raw score $a$ for worker $i$ at time step $t$ as follows:
\begin{align}\label{eq:weightedrawscore}
    a^i_t &= c_{p-1}(u^i_{t-p-1}-u^i_{t-p-2})+...+c_{1}(u^i_{t-1}-u^i_{t-2})\\
    &+c_{0}(u^i_{t}-u^i_{t-1}),
\end{align}
where $c_{p-1}+...+c_{2}+c_{1}+c_{0} = 1$. Preferably, we want to apply larger weights on the most recent terms. Then, we apply a piece-wise linear mapping to obtain our weight as described next.

For simplicity, we drop the super and subscripts for $a$ as given in equation (\ref{eq:weightedrawscore}). The following pair of $h_1$, $h_2$ are one choice, though we note that we do not have to limit these to linear functions. Let \( h_1(a) \) be a piece-wise linear function defined as follows:

\[
h_1(a) =
\begin{cases} 
1 & \text{if } a < k, \\
1 + \frac{1-\alpha}{k}(a - k) & \text{if } k \leq a \leq 0, \\
\alpha  & \text{if } 0 < a,
\end{cases}
\]
for some constant $k<0$. Similarly, we define $h_2(a)$ as:
\[
h_2(a) =
\begin{cases} 
0 & \text{if } a < k, \\
-\frac{\alpha}{k}a+\alpha & \text{if } k \leq a \leq 0, \\
\alpha & \text{if } 0 < a. 
\end{cases}
\]We replace equations (\ref{eq:worker_update}) and (\ref{eq:master_update}) with 
\begin{equation}
\theta_{t+1}^i = \theta_t^i  - h_1(a) (\theta_t^i - \theta_t^m),
\end{equation}
\begin{equation}
\theta_{t+1}^m = \theta_t^m + h_2(a) (\theta_t^i - \theta_t^m).
\end{equation}

\section{Experiment Settings}
We run our experiments on MNIST dataset\cite{lecun1998mnist} with $k \in \{4,8\}$ local workers 
and the communication period\footnote{The communication period $\tau$ controls how often a worker node communicates with the master node. For example, if $\tau = 2$, every 2 iterations, a worker will communicate with the master node.} $\tau \in \{1,2,4\}$. Our experiments are conducted 
on a single device to simulate a master-worker distributed system due to a hardware limitation. For all methods, we compare training loss and test accuracy after a full communication round and we suppress the communication between a worker node and the master node one-third of time. We use a simple 2-layer convolutional neural network from PyTorch. We compare the following methods in the results section:
\begin{itemize}
    \item \textbf{EASGD}: asynchronous version of \textbf{EASGD};
    \item \textbf{EAMSGD}: \textbf{EASGD} with \textbf{Momentum};
    \item \textbf{EAHES}: asynchronous elastic averaging \textbf{AdaHessian};
    \item \textbf{EAHES-O}: \textbf{EAHES} with data overlap (denoted as \textbf{O});
    \item \textbf{EAHES-OM}: \textbf{EAHES-O} but we manually adjust the weight $\alpha$ as if we know when a node will fail; and
    \item \textbf{DEAHES-O}: \textbf{EAHES-O} with dynamic weighting.
\end{itemize}
In most cases, it is difficult to anticipate when and which worker node will fail, However, in our experiments we will make this assumption for comparative purposes.

\section{Results}

For basic \textbf{EASGD}, \textbf{EAHES} and \textbf{EAHES-O}, we conduct a grid search on the fixed weight $\alpha$ and present the result with the best choice $\alpha =0.1$. For the data overlap approaches, we choose the ratio $r=12.5\%$ with 8 workers and $r=25\%$ with 4 workers. For \textbf{EAHES-OD}, our algorithm adjusts the weight $\alpha$ dynamically based on the change in estimated model discrepancy. The parameters for SGD-based methods are learning rate $\eta = 0.01$ and momentum $\delta = 0.5$. The parameters for AdaHessian are learning rate $\eta = 0.01$ and  $\beta = (0.9,0.999)$, and the number of sampling for Hutchinson's method is 1. We explore different choices of communication period $\tau \in \{1,2,4\}$. 

In figures \ref{fig:accuracy} and \ref{fig:loss}, due to the precise curvature information utilized by second-order methods, AdaHessian-based approaches significantly outperform those based on SGD. We observe that \textbf{EAHES-OM} has the best performance since we know when a worker will fail and can respond accordingly. \textbf{DEAHES-O} has close performance as \textbf{EAHES-OM} and outperforms all other methods. \textbf{DEAHES-O} benefits from the dynamic weight adjustment to mitigate the bad influence on the aggregated model from some potential failure without implementing extra low-level mechanisms to detect node failure. \textbf{EAHES-O} shows a better performance than 
\textbf{EAHES} indicating the positive effect of the data overlap strategy for methods associated with Hessian approximation. We also observe that as we increase the number of workers from 4 to 8 and increase communication period from 1 to 2 to 4, the performance does not degrade.

\section{Conclusions and Future Work}

In this paper, we investigate the issue of failed worker nodes in distributed deep learning and their effect on the convergence of the optimizer. We combine an elastic averaging technique with the AdaHessian optimizer and explore the effectiveness of data overlap and dynamic weighting techniques to address challenges under unfavorable conditions. Our experiments demonstrate that the proposed method significantly improves robustness and performance, effectively mitigating the impact of worker failures and optimizing training efficiency in distributed deep learning environments.

For the current report, performance under wall-clock time is not yet available. Communication rounds might not reflect the true wall-clock time due to contention among workers. Indeed, more workers will inevitably suffer from diminishing marginal utility. Our future work involves reporting results after conducting experiments on a realistic distributed system. We also aim to test our approach on practical applications employing distributed deep learning such as\cite{millard2024koopman}.
\bibliographystyle{IEEEtran}
\bibliography{references}
\end{document}